\documentclass{article}
\usepackage{spconf,amsmath,graphicx,hyperref}

\usepackage{cite}
\usepackage{amsmath,amssymb,amsfonts}
\usepackage{algorithmic}
\usepackage{graphicx}
\usepackage{textcomp}
\usepackage{xcolor}

\usepackage{wrapfig}
\usepackage{pifont}

\usepackage{float}
\usepackage{url}
\usepackage{multicol}
\usepackage{multirow}
\usepackage{hyperref}
\usepackage{colortbl} 
\definecolor{lightgray}{gray}{0.9}

\title{UniSTFormer: Unified Spatio-Temporal Lightweight Transformer for Efficient Skeleton-Based Action Recognition}

\name{Wenhan Wu$^{1}$ \quad Zhishuai Guo$^{2}$ \quad Chen Chen$^{3}$ \quad Aidong Lu$^{1}$}
\address{
$^{1}$Department of Computer Science, University of North Carolina at Charlotte, USA \\
$^{2}$Department of Computer Science, Northern Illinois University, USA \\
$^{3}$Center for Research in Computer Vision, University of Central Florida, USA \\
\texttt{\{wwu25, aidong.lu\}@uncc.edu}, \texttt{zguo@niu.edu}, \texttt{chen.chen@crcv.ucf.edu}
}

\begin{document}
\maketitle

\begin{abstract}
Skeleton-based action recognition (SAR) has achieved impressive progress with transformer architectures. However, existing methods often rely on complex module compositions and heavy designs, leading to increased parameter counts, high computational costs, and limited scalability. In this paper, we propose a unified spatio-temporal lightweight transformer framework that integrates spatial and temporal modeling within a single attention module, eliminating the need for separate temporal modeling blocks. This approach reduces redundant computations while preserving temporal awareness within the spatial modeling process. Furthermore, we introduce a simplified multi-scale pooling fusion module that combines local and global pooling pathways to enhance the model’s ability to capture fine-grained local movements and overarching global motion patterns. Extensive experiments on benchmark datasets demonstrate that our lightweight model achieves a superior balance between accuracy and efficiency, reducing parameter complexity by over 58\% and lowering computational cost by over 60\% compared to state-of-the-art transformer-based baselines, while maintaining competitive recognition performance. Our codes will be publicly available \textcolor{magenta}{\href{https://github.com/wenhanwu95/FreqMixFormer/tree/main/UniSTFormer}{here}}.

\end{abstract}
%

\begin{figure}[!t]
\vspace{-15pt}
  \centering
  \includegraphics[width=0.95\linewidth]{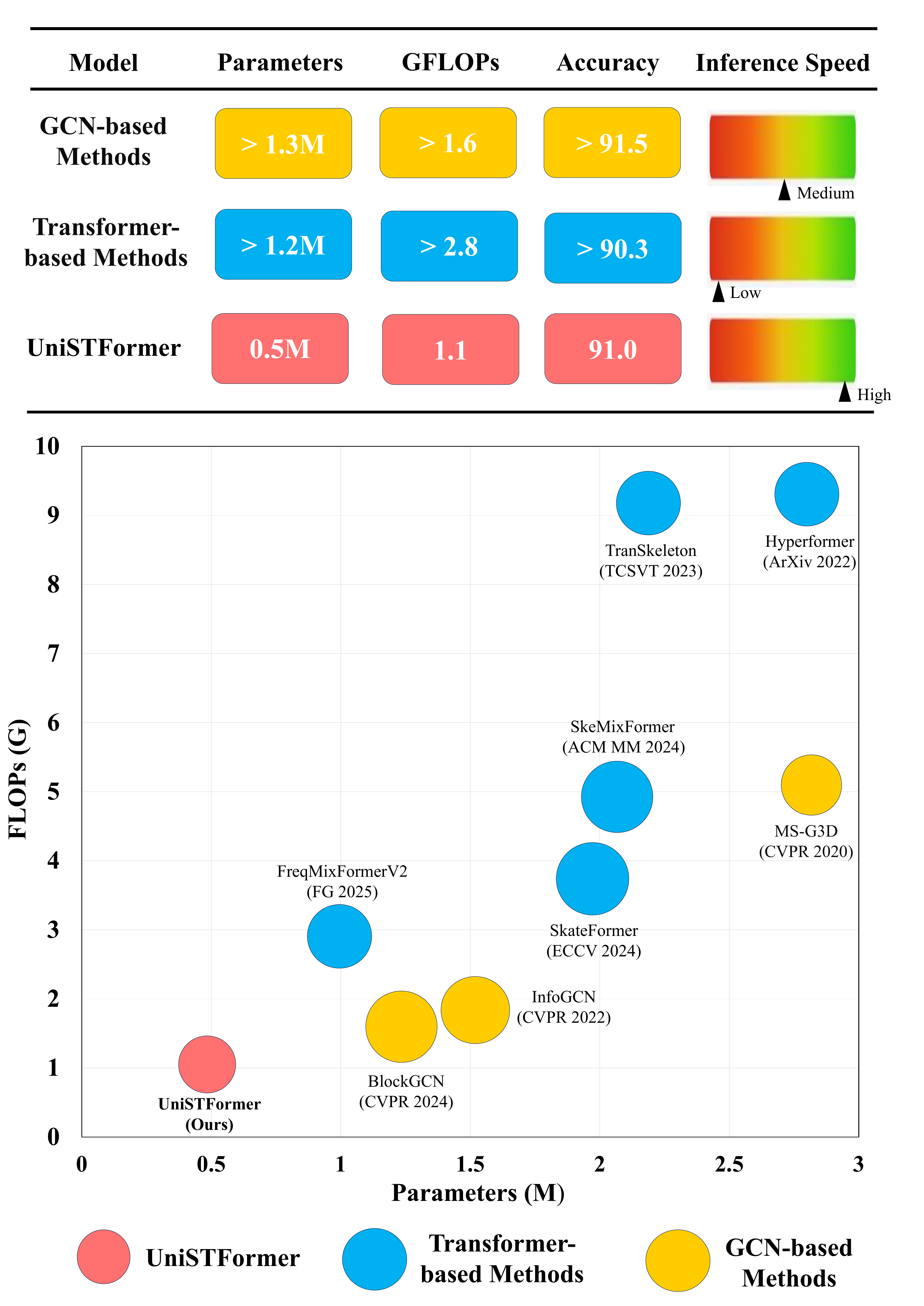}
\caption{
\textit{Top}: Comparison of model performance, size, and efficiency on the NTU-60~\cite{shahroudy2016ntu} X-Sub benchmark. UniSTFormer achieves the best trade-off among accuracy, parameter size, and computational cost.  
\textit{Bottom}: Comparison with recent SOTA methods. UniSTFormer reduces both parameters and GFLOPs by over 58\% while maintaining competitive recognition performance (91.0\%). \textit{The marker size reflects classification accuracy.} 
For more detailed comparisons, please refer to Table~\ref{tab: results} and Table~\ref{tab:efficiency}.
}
  \label{fig:fig1}
  \vspace{-15pt}
\end{figure}

\section{Introduction}
Skeleton-based action recognition (SAR)  \cite{liu2020disentangling, chen2021channel, chi2022infogcn, plizzari2021skeleton, xin2023skeleton, wu2024frequency} has emerged as a fundamental task in computer vision. Transformer-based architectures \cite{plizzari2021skeleton, zhou2022hypergraph, liu2023transkeleton, xin2023skeleton, wu2024frequency} have played a significant role in driving the remarkable progress of SAR in recent years. These models capture the complex spatial and temporal dependencies inherent in human skeleton sequences, modeling the dynamic evolution of joint movements over time and the intricate structural relationships among body parts.

Despite recent progress, current transformer-based SAR models still face several fundamental limitations that hinder both their efficiency and generalization capacity. In particular, many methods adopt a two-branch design that separately models temporal and spatial dependencies, \textbf{typically by stacking or aggregating temporal and spatial modules redundantly} (\textcolor{blue}{\textbf{Challenge 1}}). For example, ST-TR~\cite{plizzari2021skeleton} employs sequential spatial-temporal attention blocks, while Hyperformer~\cite{zhou2022hypergraph} integrates hypergraph-based transformer layers to recombine multi-scale dependencies. Although these designs effectively capture dynamic and structural patterns, they inevitably introduce duplicated feature transformations across branches, leading to redundant computations, inflated parameter counts, and increased architectural complexity. As model complexity increases, the interaction between temporal and spatial blocks further complicates optimization and slows inference, limiting their practical deployment.

Moreover, many existing transformer approaches adopt overly complex mechanisms to integrate global and local features, often \textbf{relying on heuristic-based or pre-set fusion strategies} (\textcolor{blue}{\textbf{Challenge 2}}). For example, SkeMixFormer~\cite{xin2023skeleton} applies a multivariate topology design to separately process different spatial relations before fusing them, while FreqMixFormer~\cite{wu2024frequency} introduces frequency-domain modulation but employs a dense combination of spatial and frequency attention blocks. While these methods enhance local-global interactions, they also increase computational overhead and restrict scalability. This underscores the need for lightweight yet powerful SAR frameworks that efficiently capture multi-scale spatio-temporal dependencies while reducing redundancy and computational costs.

In this work, we introduce a unified lightweight spatio-temporal transformer framework that directly addresses these limitations. \textcolor{blue}{\textbf{To address Challenge 1}}, we design a unified attention mechanism that applies joint-wise attention to each frame, thereby integrating spatial and temporal modeling in a single block without explicit temporal modules. This allows our network to preserve full temporal correlations and capture dynamic motion patterns through shared attention and channel refinement. Moreover, we propose a multi-scale pooling attention module that combines global average pooling and adaptive local pooling to generate joint-wise attention maps, enabling the network to adaptively capture both global motion context and subtle local joint dynamics without relying on complex, handcrafted fusion designs (\textcolor{blue}{\textbf{Tackling Challenge 2}}). Through extensive ablation studies and experiments on standard SAR benchmarks, we demonstrate that our approach achieves a superior balance between efficiency and accuracy, outperforming state-of-the-art methods while maintaining a significantly reduced parameter footprint (see Figure~\ref{fig:fig1}). 

Our key contributions are: 
\textbf{1) }We propose a unified module that integrates temporal and spatial modeling in a single attention mechanism, reducing redundancy and simplifying the architecture.
\textbf{2) }We introduce a multi-scale pooling attention that fuses global and local features, achieving a better trade-off between capturing coarse and fine-grained actions and maintaining a significantly reduced parameter footprint.
\textbf{3)} Extensive experiments demonstrate that our design achieves competitive accuracy while reducing parameters by over \textbf{58\%} and computational cost by over \textbf{60\%} compared to SOTA models.

\begin{figure*}[t]
\vspace{-15pt}
  \centering
  \includegraphics[width= 0.95\linewidth]{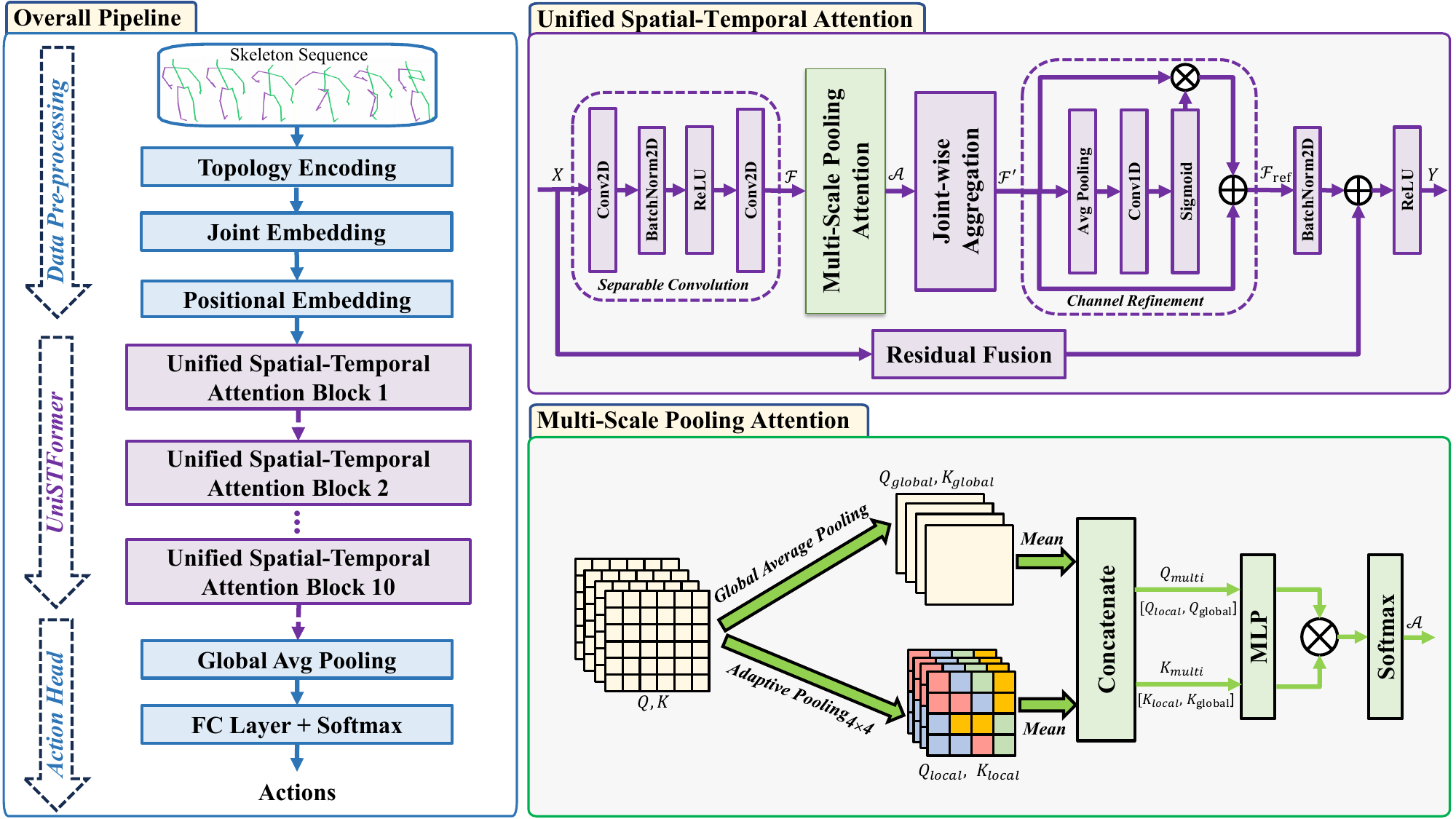}
  \vspace{-10pt}
    \caption{
    Overview of the proposed UniSTFormer framework. \textit{Left}: The model consists of 10 stacked unified spatial-temporal attention blocks built upon topology-aware skeleton encoding and joint embedding. \textit{Top-right}: Each block integrates separable convolution, multi-scale attention, and channel refinement with residual fusion (Section~\ref{sec: unified}). \textit{Bottom-right}: The multi-scale pooling module combines global and local joint representations to compute dynamic attention maps (Section~\ref{sec: multi}).
    }
  \label{fig:fig2}
  \vspace{-15pt}
\end{figure*}

\section{Method}
\subsection{Unified Spatial-Temporal Attention}
\label{sec: unified}
Given an input skeleton sequence \(X \in \mathbb{R}^{N \times C \times T \times V}\), where \(N\) is the batch size, \(C\) is the number of input channels (e.g., 3 for \((x, y, z)\)), \(T\) is the number of frames, and \(V\) is the number of joints, our goal is to predict the action class label \(y \in \{1, \ldots, K\}\). 
To achieve efficient yet expressive modeling, we propose a unified spatial attention mechanism that eliminates the need for explicit temporal modeling blocks. Specifically, the input \(X\) is first transformed using a separable convolution, which reduces parameter overhead while preserving spatial structure:
{\setlength{\abovedisplayskip}{4pt}
 \setlength{\belowdisplayskip}{4pt}
\begin{equation}
\mathcal{F} = \text{Conv}_{\text{sep}}(X), \quad \mathcal{F} \in \mathbb{R}^{N \times C' \times T \times V},
\label{eq:convsep}
\end{equation}
}
where \(C'\) is the transformed feature dimension.

We then compute a dynamic attention map \(\mathcal{A} \in \mathbb{R}^{N \times V \times V}\) from \(\mathcal{F}\) using a multi-scale pooling attention module (see Section~\ref{sec: multi}), which captures both global context and local motion features. This dynamic map is fused with a learnable static topology prior \(\mathcal{A}_{\text{init}}\) to form the final attention matrix \(\mathcal{M}\), which enhances joint-wise spatial dependencies in a topology-aware manner:
{\setlength{\abovedisplayskip}{4pt}
 \setlength{\belowdisplayskip}{4pt}
\begin{equation}
\mathcal{M} = \alpha \cdot \mathcal{A} + (1 - \alpha) \cdot \mathcal{A}_{\text{init}},
\end{equation}
}
where \(\alpha\) is a learnable scalar. The prior \(\mathcal{A}_{\text{init}}\) is initialized from the binary adjacency matrix of the skeleton graph, where two joints are connected if there exists a direct kinematic link between them. \(\mathcal{M}\) is then applied to each time step \(t\) of the feature map \(\mathcal{F}\), denoted as \(\mathcal{F}_t\):
{\setlength{\abovedisplayskip}{4pt}
 \setlength{\belowdisplayskip}{4pt}
\begin{equation}
\mathcal{F}'_t = \mathcal{F}_t \mathcal{M}^\top, \quad \forall t.
\label{eq:attnapply}
\end{equation}
}
To further enhance the channel-wise discriminability, we apply a lightweight yet effective channel refinement strategy inspired by \cite{xin2023skeleton, hu2018squeeze}. Specifically, we first perform global average pooling to aggregate feature statistics along the temporal and joint dimensions, followed by a 1D convolution to model inter-channel dependencies. A sigmoid gating function is then applied to generate channel-wise attention weights, which are used to rescale the original features in a residual manner:
{\setlength{\abovedisplayskip}{4pt}
 \setlength{\belowdisplayskip}{4pt}
\begin{equation}
\mathcal{F}_{\text{ref}} = \mathcal{F}^{\prime} + \mathcal{F}^{\prime} \cdot \sigma(\text{Conv1D}(\text{AvgPool}(\mathcal{F}^{\prime}))).
\end{equation}
}
The final output feature is obtained by adding a residual connection \(\mathbf{R}(X)\) and applying batch normalization and ReLU activation:
{\setlength{\abovedisplayskip}{4pt}
 \setlength{\belowdisplayskip}{4pt}
\begin{equation}
\textit{Y} = \text{ReLU}(\text{BN}(\mathcal{F}_{\text{ref}}) + \mathbf{R}(X)).
\end{equation}
}

At the end of the network, global average pooling (GAP) over temporal and joint dimensions is applied, followed by a linear classifier for action recognition:
{\setlength{\abovedisplayskip}{4pt}
 \setlength{\belowdisplayskip}{4pt}
\begin{equation}
y = \text{Softmax}(\text{FC}(\text{GAP}_{T,V}(\textit{Y}))).
\end{equation}
}

This unified attention block (see Figure~\ref{fig:fig2}) integrates spatial and temporal modeling by applying joint-wise attention to each frame while preserving the full temporal resolution. Similar to \cite{xin2023skeleton}, our network stacks 10 unified attention blocks sequentially for deep representation learning.

Although no explicit temporal modules are used, temporal evolution emerges implicitly. The dynamic attention map \(\mathcal{A}\) first encodes global temporal context by pooling features across the entire time dimension (Eq.~\ref{eq:convsep} - ~\ref{eq:attnapply} and Section~\ref{sec: multi}), allowing temporal information to be preserved across the sequence. Subsequently, stacking these unified blocks enables the model to refine these temporal dependencies progressively. This unified design improves computational efficiency by modeling spatio-temporal patterns without dedicated temporal modules \textcolor{blue}{(addressing Challenge 1)}.

\vspace{-15pt}
\subsection{Multi-Scale Pooling Attention}
\label{sec: multi}
Given an input feature map \(\mathcal{F} \in \mathbb{R}^{N \times C \times T \times V}\), we divide the channels to compute query (\(Q\)) and key (\(K\)):
{\setlength{\abovedisplayskip}{4pt}
 \setlength{\belowdisplayskip}{4pt}
\begin{equation}
Q = \mathcal{F}[:, :C/2, :, :], \quad K = \mathcal{F}[:, C/2:, :, :].
\end{equation}
}

Each branch is processed using two spatial pooling operations to capture both global context and local motion features:
{\setlength{\abovedisplayskip}{4pt}
 \setlength{\belowdisplayskip}{4pt}
\begin{align}
Q_{\text{global}} &= \text{AvgPool}_{C,T}(Q) \in \mathbb{R}^{N \times V}, \\
Q_{\text{local}}  &= \text{Mean}(\text{AdaptivePool}_{4\times4}(Q)) \in \mathbb{R}^{N \times V}, \\
K_{\text{global}} &= \text{AvgPool}_{C,T}(K) \in \mathbb{R}^{N \times V}, \\
K_{\text{local}}  &= \text{Mean}(\text{AdaptivePool}_{4\times4}(K)) \in \mathbb{R}^{N \times V}.
\end{align}
}

These pooled representations are concatenated to form multi-scale joint representations:
{\setlength{\abovedisplayskip}{4pt}
 \setlength{\belowdisplayskip}{4pt}
\begin{align}
Q_{\text{multi}} &= [Q_{\text{global}}; Q_{\text{local}}] \in \mathbb{R}^{N \times 2V}, \\
K_{\text{multi}} &= [K_{\text{global}}; K_{\text{local}}] \in \mathbb{R}^{N \times 2V}.
\end{align}
}


We project both through a multi-layer perceptron (MLP) with ReLU and dropout. The attention matrix $\mathcal{A} \in \mathbb{R}^{N \times V \times V}$ is then computed via a batched outer product followed by softmax along the last dimension:
{\setlength{\abovedisplayskip}{4pt}
 \setlength{\belowdisplayskip}{4pt}
\begin{equation}
\mathcal{A} = \text{Softmax}\left( \text{MLP}(Q_{\text{multi}}) \otimes \text{MLP}(K_{\text{multi}}) \right).
\label{eq:final_attention}
\end{equation}
}
This attention map simplifies the fusion of global and local features by leveraging global and local pooling. It enables the network to focus on joints at multiple scales adaptively. This design also avoids hand-crafted fusion or complex block stacking, enabling efficient local-global integration with minimal computation \textcolor{blue}{(addressing Challenge 2)}.

\begin{table*}[t]
\scriptsize
\renewcommand\arraystretch{1.0}
\centering
  \caption{Comparison with the SOTA.}
  \setlength\tabcolsep{6.5pt}
{
\begin{tabular}{c|c|c|c|c|c|c|c|c|c}
\hline
\multirow{2}{*}{Methods} & \multirow{2}{*}{Venue} & \multirow{2}{*}{Category} & \multirow{2}{*}{Params (M)} & \multirow{2}{*}{FLOPs (G)} & \multicolumn{2}{c|}{NTU-60} & \multicolumn{2}{c|}{NTU-120} & \multirow{2}{*}{NW-UCLA} \\ \cline{6-9}
 & & & & & X-Sub (\%) & X-View (\%) & X-Sub (\%) & X-Set (\%) & \\ \hline
MS-G3D \cite{liu2020disentangling} & CVPR'2020 & GCN & 2.8 (\textcolor{red}{$\downarrow$ 82.1\%}) & 5.2 (\textcolor{red}{$\downarrow$ 78.8\%}) & 91.5 & 96.2 & 86.9 & 88.4 & - \\
CTR-GCN \cite{chen2021channel} & ICCV'2021 & GCN & 1.5 (\textcolor{red}{$\downarrow$ 66.7\%}) & 2.0 (\textcolor{red}{$\downarrow$ 45.0\%}) & 92.4 & 96.4 & 88.9 & 90.4 & 96.5 \\
EfficientGCN-B4 \cite{song2022constructing} & TPAMI'2022 & GCN & 2.0 (\textcolor{red}{$\downarrow$ 75.0\%}) & 15.2 (\textcolor{red}{$\downarrow$ 92.8\%}) & 91.7 & 95.7 & 88.3 & 89.1 & - \\
InfoGCN \cite{chi2022infogcn} & CVPR'2022 & GCN & 1.6 (\textcolor{red}{$\downarrow$ 68.8\%}) & 1.8 (\textcolor{red}{$\downarrow$ 38.9\%}) & 92.3 & 96.7 & 89.2 & 90.7 & 96.6 \\
FRHead \cite{zhou2023learning} & CVPR'2023 & GCN & 2.0 (\textcolor{red}{$\downarrow$ 75.0\%}) & - & 92.8 & 96.8 & 89.5 & 90.9 & 96.8 \\
BlockGCN \cite{zhou2024blockgcn} & CVPR'2024 & GCN & 1.3 (\textcolor{red}{$\downarrow$ 61.5\%}) & 1.6 (\textcolor{red}{$\downarrow$ 31.3\%}) & 93.1 & 97.0 & 90.3 & 91.5 & 96.9 \\ 
DeGCN \cite{myung2024degcn} & TIP'2024 & GCN & 5.6 (\textcolor{red}{$\downarrow$ 91.0\%}) & - & 93.6 & 97.4 & 91.0 & 92.1 & 97.2 \\ \hline
ST-TR \cite{plizzari2021skeleton} & CVIU'21 & Transformer & 12.1 (\textcolor{red}{$\downarrow$ 95.9\%}) & 259.4 (\textcolor{red}{$\downarrow$ 99.6\%}) & 90.3 & 96.3 & 85.1 & 87.1 & - \\
TranSkeleton \cite{liu2023transkeleton} & TCSVT'23 & Transformer & 2.2 (\textcolor{red}{$\downarrow$ 77.3\%}) & 9.2 (\textcolor{red}{$\downarrow$ 88.0\%}) & 92.8 & 97.0 & 89.4 & 90.5 & - \\
Hyperformer \cite{zhou2022hypergraph} & arXiv'2022 & Transformer & 2.7 (\textcolor{red}{$\downarrow$ 81.5\%}) & 9.6 (\textcolor{red}{$\downarrow$ 88.5\%}) & 92.9 & 96.5 & 89.9 & 91.3 & 96.7 \\
SkeMixFormer \cite{xin2023skeleton} & ACMM'2023 & Transformer & 2.1 (\textcolor{red}{$\downarrow$ 76.2\%}) & 4.8 (\textcolor{red}{$\downarrow$ 77.1\%}) & 93.0 & 97.1 & 90.1 & 91.3 & 97.4 \\
SkateFormer \cite{do2025skateformer} & ECCV'2024 & Transformer & 2.0 (\textcolor{red}{$\downarrow$ 75.0\%}) & 3.6 (\textcolor{red}{$\downarrow$ 69.4\%}) & 93.5 & 97.4 & 89.8 & 91.4 & 98.3 \\
FreqMixFormer \cite{wu2024frequency} & ACMM'2024 & Transformer & 2.0 (\textcolor{red}{$\downarrow$ 75.0\%}) & 64.4 (\textcolor{red}{$\downarrow$ 98.3\%}) & 93.6 & 97.4 & 90.5 & 91.9 & 97.4 \\ 
FreqMixFormerV2 \cite{wu2024freqmixformerv2} & FG'2025 & Transformer & 1.2 (\textcolor{red}{$\downarrow$ 58.3\%}) & 2.8 (\textcolor{red}{$\downarrow$ 60.7\%}) & 92.9 & 96.9 & 90.0 & 91.1 & 97.0 \\ \hline
\rowcolor{gray!30} \textbf{Ours} & - & Transformer & \textbf{\textcolor{red}{0.5}} & \textbf{\textcolor{red}{1.1}} & 91.0 & 95.3 & 86.9 & 88.2 & 96.1 \\ \hline
\end{tabular}
}
\label{tab: results}
\vspace{-15pt}
\end{table*}
\vspace{-15pt}

\section{Experiments}
\subsection{Datasets and Implementation}
We evaluate on NTU-60~\cite{shahroudy2016ntu}, NTU-120~\cite{liu2019ntu}, and NW-UCLA~\cite{wang2014cross} datasets under standard protocols. NTU provides 3D skeletons with 25 joints; NW-UCLA offers multi-view Kinect sequences. Our implementation follows \cite{xin2023skeleton}, using the same pre-processing and training pipeline. We use 64-frame sequences, batch size 128, and train for 100 epochs with SGD (LR=0.1, weight decay=0.0005). All experiments are run on a single RTX A6000. Ensemble strategy follows~\cite{chi2022infogcn, xin2023skeleton, wu2024frequency}, and final (6-ensemble) results are shown in Table~\ref{tab: results}.
\subsection{Comparison With the State-of-the-Art}
As shown in Table~\ref{tab: results}, our method achieves a compelling balance between accuracy and efficiency. It obtains 91.0\% accuracy on NTU-60 X-Sub with only 0.5M parameters and 1.1 GFLOPs, which is significantly lower than SkeMixFormer\cite{xin2023skeleton} (2.1M, 4.8 GFLOPs) and FreqMixFormerV2 \cite{wu2024freqmixformerv2}(1.2M, 2.8 GFLOPs). Compared to GCN-based methods such as BlockGCN~\cite{zhou2024blockgcn} (93.1\%, 1.3M, 1.6 GFLOPs), our ensemble achieves similar performance with over 60\% fewer parameters and 30\% less computation cost. These results validate the effectiveness of our unified attention design in capturing spatial-temporal patterns with minimal overhead.

\vspace{-10pt}
\subsection{Ablation Study}
We analyze the impact of the hidden dimension in the MLP for attention computation to identify the trade-off between representational capacity and efficiency (NTU-60 X-Sub settings). As shown in Table~\ref{tab: MLP}, increasing the hidden size from 32 to 128 consistently improves performance, achieving the best accuracy (86.2\%) at 128. Increasing to 256 causes a marginal drop and adds parameter overhead (0.6M). Therefore, we adopt 128 as a balanced setting between accuracy and model size to preserve effectiveness and compactness. We also assess the multi-scale pooling attention by comparing global pooling, local pooling, and their combination. As shown in Table~\ref{tab: pooling}, global and local pooling achieve 83.9\% and 84.7\%, respectively, while their combination boosts accuracy to 86.2\% with minimal overhead (0.54M vs. 0.51M). This confirms the complementary benefits of global context and localized patterns, highlighting the importance of multi-scale design for enhanced spatial sensitivity at low cost.

\begin{table}[t]
\centering
\begin{minipage}[t]{0.47\linewidth}
\centering
\caption{MLP hidden dimension vs. accuracy.}
\label{tab:mlp_hidden}
\vspace{5pt}
\scriptsize
\begin{tabular}{@{}c|c|c@{}}
\hline
\textbf{Dim} & \textbf{Params} & \textbf{Acc.} \\
\hline
32  & 0.4M & 82.5 \\
64  & 0.4M & 84.3 \\
128 & 0.5M & \textbf{86.2} \\
256 & 0.6M & 84.9 \\
\hline
\end{tabular}
\label{tab: MLP}
\end{minipage}
\vspace{-15pt}
\hfill
\begin{minipage}[t]{0.47\linewidth}
\centering
\caption{Multi-scale attention variants.}
\label{tab:multi_attn}
\vspace{10pt}
\scriptsize
\begin{tabular}{@{}c|c|c@{}}
\hline
\textbf{Variant} & \textbf{Params} & \textbf{Acc.} \\
\hline
Global Only     & 0.51M & 83.9 \\
Local Only      & 0.51M & 84.7 \\
Global + Local  & 0.54M & \textbf{86.2} \\
\hline
\end{tabular}
\label{tab: pooling}
\end{minipage}
\end{table}
\vspace{-20pt}

\begin{table}[h]
\centering
\caption{Efficiency comparison with the released codes. NTU-60 X-Sub setting.}
\renewcommand{\arraystretch}{1.2}
\small
\resizebox{\columnwidth}{!}{
\begin{tabular}{lccccc}
\hline
\textbf{Method} & \textbf{Type} & \textbf{Params (M)} & \textbf{FLOPs (G)} & \textbf{Inference Time (ms)} & \textbf{Accuracy (\%)} \\
\hline
CTR-GCN         & GCN         & 1.6 (\textcolor{red}{$\downarrow$ 68.8\%}) & 2.0 (\textcolor{red}{$\downarrow$ 45.0\%}) & 24.2 (\textcolor{red}{$\downarrow$ 53.7\%}) & 92.4 \\
BlockGCN        & GCN         & 1.3 (\textcolor{red}{$\downarrow$ 61.5\%}) & 1.6 (\textcolor{red}{$\downarrow$ 31.3\%}) & 19.0 (\textcolor{red}{$\downarrow$ 41.1\%}) & 93.1 \\
\hline
SkeMixFormer    & Transformer & 2.1 (\textcolor{red}{$\downarrow$ 76.2\%}) & 4.8 (\textcolor{red}{$\downarrow$ 77.1\%}) & 48.2 (\textcolor{red}{$\downarrow$ 76.8\%}) & 93.0 \\
FreqMixFormer   & Transformer & 2.0 (\textcolor{red}{$\downarrow$ 75.0\%}) & 64.4 (\textcolor{red}{$\downarrow$ 98.3\%}) & 199.1 (\textcolor{red}{$\downarrow$ 94.4\%}) & 93.6 \\
FreqMixFormerV2 & Transformer & 1.2 (\textcolor{red}{$\downarrow$ 58.3\%}) & 2.8 (\textcolor{red}{$\downarrow$ 60.7\%}) & 82.0 (\textcolor{red}{$\downarrow$ 86.3\%}) & 92.9 \\
\hline
\textbf{Ours}   & Transformer & \textbf{0.5} & \textbf{1.1} & \textbf{11.2} & 91.0 \\
\hline
\end{tabular}
}
\vspace{-15pt}
\label{tab:efficiency}
\end{table}

\subsection{Efficiency Analysis}
Table~\ref{tab:efficiency} provides a comprehensive comparison of model efficiency with representative transformer-based and GCN-based models, including parameters, computational cost (GFLOPs), and inference time (for a fair comparison, we set the input sequence length $T=64$ for all models, and all the experiments are tested on the same hardware). As shown in Table~\ref{tab:efficiency}, our method achieves remarkable efficiency improvements compared to prior transformer-based approaches. It reduces parameter count to 0.5M and GFLOPs to 1.1, which is a substantial drop of over 76\% and 77\% respectively compared to SkeMixFormer\cite{xin2023skeleton}. Notably, our inference time is only 11.2 ms—less than half of SkeMixFormer (48.2 ms) and nearly 7.5$\times$ faster than FreqMixFormerV2 (82.0 ms). Despite the efficiency gains, our model maintains competitive accuracy (91.0\%), demonstrating that our unified attention design achieves a superior balance between inference speed, model size, and recognition accuracy.

\section{Conclusion}
This paper proposes UniSTFormer, a unified and lightweight spatio-temporal transformer for skeleton-based action recognition. By integrating spatial and temporal modeling into a unified attention mechanism, our design eliminates the need for explicit temporal convolutions while maintaining temporal awareness. We further employ a multi-scale pooling module to effectively capture both local and global motion patterns. Extensive experiments demonstrate that our method achieves a remarkable balance between accuracy and efficiency, outperforming prior lightweight models with fewer parameters, lower GFLOPs, and faster inference.


\bibliographystyle{IEEEbib}
\bibliography{strings,refs}

\end{document}